\documentclass{article}
\usepackage{arxiv}

\usepackage[utf8]{inputenc} 
\usepackage[T1]{fontenc}    
\usepackage{hyperref}       
\usepackage{url}            
\usepackage{booktabs}       
\usepackage{amsfonts}       
\usepackage{nicefrac}       
\usepackage{microtype}      

\usepackage{array}
\usepackage{textcomp}
\usepackage{stfloats}
\usepackage{url}
\usepackage{verbatim}
\usepackage{graphicx,setspace,lscape,longtable,epsfig}
\usepackage{amsmath,amsthm,amssymb,amsfonts}
\usepackage{color,xcolor}
\usepackage{hyperref}
\usepackage{float}
\usepackage{mdframed,lipsum}
\usepackage[sort, numbers]{natbib}
\usepackage{algorithmic,algorithm}
\usepackage[caption=false,font=normalsize,labelfont=sf,textfont=sf]{subfig}
\numberwithin{equation}{section}  

\newtheorem{thm}{Theorem}[section]

\newtheorem{prop}{Proposition}[section]

\usepackage{xr}
\makeatletter

\usepackage{mathtools}
\usepackage{mathrsfs}
\usepackage{multirow}
\usepackage{cite}
\usepackage{booktabs} 
\usepackage{pifont}
\makeatletter

\def\bec{\begin{center}}
\def\enc{\end{center}}
\numberwithin{equation}{section}

\begin{document}
\include{math_cmds}
\title{A Two-Sample Test of Text Generation Similarity}
\author{Jingbin Xu, Chen Qian,
        Meimei Liu, Feng Guo
\thanks{
Jingbin Xu is with School of Mechanical Engineering, Dalian University of Technology, Dalian, China; Chen Qian is with School of Economics and Management,  Dalian University of Technology, Dalian, China; Meimei Liu and Feng Guo are with the Department of Statistics, Virginia  Tech,  Blacksburg, VA, USA.}
}
\maketitle

\begin{abstract}
   The surge in digitized text data requires reliable {inferential} methods on observed textual patterns. This article proposes a {novel} two-sample {text} test for comparing similarity between two groups of documents. The hypothesis is whether the probabilistic mapping generating the textual data is identical across two groups of documents. The proposed test aim{s} to assess text similarity by comparing the entropy of the documents. Entropy is estimated using neural network-based language models. The test statistic is derived from an estimation-and-inference framework, where the entropy is first approximated using an estimation set, followed by inference on the remaining data set. {We showed theoretically that} under mild conditions, the test statistic’s asymptotically follows a normal distribution. A multiple data-splitting strategy is proposed to enhance test power, which combines p-values into a unified decision. Various simulation studies and a real data example {demonstrated} that the proposed two-sample text test maintains the {nominal} Type I error rate while offering greater power compared to existing methods. The proposed method provides a novel solution to assert differences in document classes, particularly in fields where large-scale textual information is crucial. 
\vspace{3mm}

\end{abstract}

\section{Introduction}
\noindent 
The explosive growth of digitized textual information brings challenges to text analysis. Determining whether two collections of documents are similar is crucial in various applications, such as identifying related electronic health records \citep{yin2020learning}, mining social media content \citep{trotzek2018utilizing, zhong2023feature}, and tracking shifts in political opinions through polls \citep{grimmer2013text}. {T}he rise of Artificial Intelligence (AI)-generated text has sparked debates {on} copyright protection between AI systems and human authors \citep{herbold2023large}. {T}ext similarity metrics {have also been used} to measure the level of information in computation linguistics \citep{kontoyiannis1998nonparametric,lin2013similarity} and to assess the performance of large language models \citep{brown2020language}. It is essential to develop reliable methods to quantify differences between groups of documents. 

Applying statistical test methods to high-dimensional data encounters challenges in text analysis. Common hypothesis testing techniques for high-dimensional data include distance correlation metrics tailored for high-dimensional random vectors \citep{gao2021asymptotic} and projection tests on mean vectors \citep{liu2022projection}. Although these methods have been shown to be effective in analyzing complex data types, such as longitudinal data with intricate correlations \citep{fang2020test}, they fail to address the unstructured nature of textual data. Textual data are inherently sparse, high-dimensional, and characterized by intricate internal logical connections \citep{grimmer2022text}. These unique features pose challenges for conducting reliable statistical inference on textual data.

Given the unstructured nature of text data \citep{wu2013data}, text analysis typically begins by converting discrete terms into continuous representations. One approach is the use of a document-term matrix, which captures the frequency of terms or words within each document \citep{sparck1972statistical}. As seen in the vector space model \citep{jing2010knowledge}, a document is treated as a ``bag" of word occurrences \citep{zhao2017fuzzy}, where the dependence structure across words or terms is omitted. However, these dependencies are closely tied to semantic meanings, which are essential for a deeper understanding of a document's content. Semantics explores how context shapes meaning and how variations in words and structures can convey nuanced interpretations \citep{chowdhary2020natural}. 

To capture this semantic information, researchers have leveraged neural network-based language models that learn from the sequence of words in text data \citep{bengio2000neural}. For instance, \textit{Word2Vec} transforms high-dimensional one-hot encoded-word matrices into continuous vector spaces through embedding learning \citep{mikolov2013distributed}. This approach captures the semantic meaning of words based on their contextual relationships within a document. Probabilistic methods are also widely used to model the text generation process. Latent Dirichlet Allocation (LDA) \citep{blei2003latent}, a Bayesian framework, identifies topics within text corpora by modeling the topic-word distribution with a Dirichlet prior. The prior reflects the intuition that word distributions within topics are often skewed, with only a small subset of words having high probabilities. 

Text similarity analysis has been gaining significant attention, ranging from simple keyword matching \citep{lovins1968development} to {advanced} vector-based metrics. {Once textual data is converted into numerical representations, various methods can be applied to assess text similarity. For instance, a Maximum Mean Discrepancy (MMD) procedure is proposed to infer whether two sets of documents convey similar meanings based on the vector space model \citep{jitkrittum2016interpretable}.} Matrix factorization techniques, such as Latent Semantic Analysis (LSA), have also been employed to construct vector-based metrics for document comparison \citep{landauer1998introduction}. Other measures, such as Kullback–Leibler (KL) divergence \citep{dhillon2003divisive} and the Wasserstein distance, have been used to evaluate differences between distributions. For instance, Word Mover's Distance (WMD) {adopted} the Wasserstein distance to assess document similarity \citep{kusner2015word}. Unlike methods that rely solely on word occurrences or frequencies, WMD captures the semantic meaning of a document by combining individual word embedding, accounting for word meanings in a high-dimensional space. Recent advancements enhance these methods using pre-trained large language models, such as Generative Pre-trained Transformers (GPT) \citep{brown2020language} for semantic meaning comparisons \citep{pillutla2021mauve}.   

Text similarity comparison faces several challenges. First, aligning word embeddings with learning objectives can be challenging, as many methods prioritize the prediction of the next word without incorporating an inference mechanism. This misalignment elevates the risk of false discoveries \citep{dai2022embedding}. Second, the computational complexity of two-stage statistical inference can be expensive, especially with a large text corpus. {In the first stage, raw text data is transformed into high-dimensional numerical representations using embedding techniques. Once the data is in numerical form, statistical inference is applied.} For instance, the MMD method applied to numerical embedding has a computational complexity of \(O(N^2V)\) \citep{yan2023kernel}, where $N$ is the number of documents and $V$ is the number of unique words. In addition, estimating the null distribution of MMD often relies on computationally expensive Monte Carlo permutation techniques \citep{jitkrittum2016interpretable}. Third, neural network-based language models are being scrutinized for their transparency and interoperability deficits \citep{dai2022significance}. The statistical examination of text similarity, particularly concerning AI-generated text, demands further exploration. 

This study addresses the aforementioned challenges using a two-sample text test that incorporates probability sample spaces for the document generation process. The method aims to compare the information content of two sets of documents, testing whether they are generated by the same probabilistic mapping measure. The test statistic utilizes neural network-based language models, shifting the analytical focus from the word level to the document level to represent the text corpus entropy. The document-level analysis enables examining the underlying probabilistic measures in texts. The asymptotic behavior of the test statistic is derived from an estimation-and-inference framework: entropy is approximated using an estimation set, followed by inference on the remaining set. Although necessary for training neural language models, the estimation and inference framework reduces the power of the test. To improve the robustness of hypothesis testing, we employ a multiple-data splitting strategy by aggregating \textit{p}-values at the inference phase. This strategy compensates for potential power loss, ensuring an effective and reliable testing process. 

The article is organized as follows. Section~\ref{sec: background on text} provides an overview of the neural network language models and the text generation process. Section~\ref{sec: two sample text test} introduces the test statistic, examines its asymptotic behaviors, and discusses the training of neural network language models. The proposed algorithm incorporates a data-splitting strategy for estimation and inference, with multiple data-splitting to mitigate power loss. Section~\ref{sec: numerical results} presents numerical studies, including Monte Carlo numerical simulations, AG News benchmark experiment, and the case study of the U.S. national crash report database. The numerical studies compare the finite sample performances of the proposed methods against competing approaches. The summary and discussion are provided in Section~\ref{sec: conclusion}. All technical proofs and additional numerical results are included in the supplementary materials. 

\section{Problem Definition}
\label{sec: background on text}
Given a dictionary set consisting of $V$ unique words, denoted as $\{v_1,...,v_V\}$. Let $D$ be a random variable representing a document, defined as a sequence of words $D = \{W_1, \ldots, W_T\}$, where $W_t$ is a random variable representing the word at position $t$, and $T$ is the document length. Each $W_t$ is drawn from the dictionary. Proposition~\ref{ProbabilitySpace} states that the random variable $D$ is generated from a probability space $(\Omega, \mathcal{F}, \mathrm{P})$ based on the dictionary set $\{v_1,...v_V\}$. 

{\prop [Probability space for document]
\label{ProbabilitySpace}  
Define an event $D_n$ as a possible combination of words from the dictionary set $\{v_1,...,v_V\}$. Define the sample space $\Omega = \cup_{n=1}^{\infty}D_n$ as the collections of all possible word sequences that can be formed from the given dictionary. Let $\mathrm{P}$ be a probabilistic measure mapping defined as $ \mathrm{P}: \Omega\rightarrow[0,1]$, satisfying the property $\sum_{n=1}^{\infty} \mathrm{P}(D_n)=1$.  The sample space $\Omega$ is infinite but countable. With $\mathcal{F}$ defined as the collection of subsets of $\Omega$, $(\Omega, \mathcal{F}, \mathrm{P})$ forms a probability space. }

\begin{figure}[ht]
\centering
\includegraphics[scale=1]{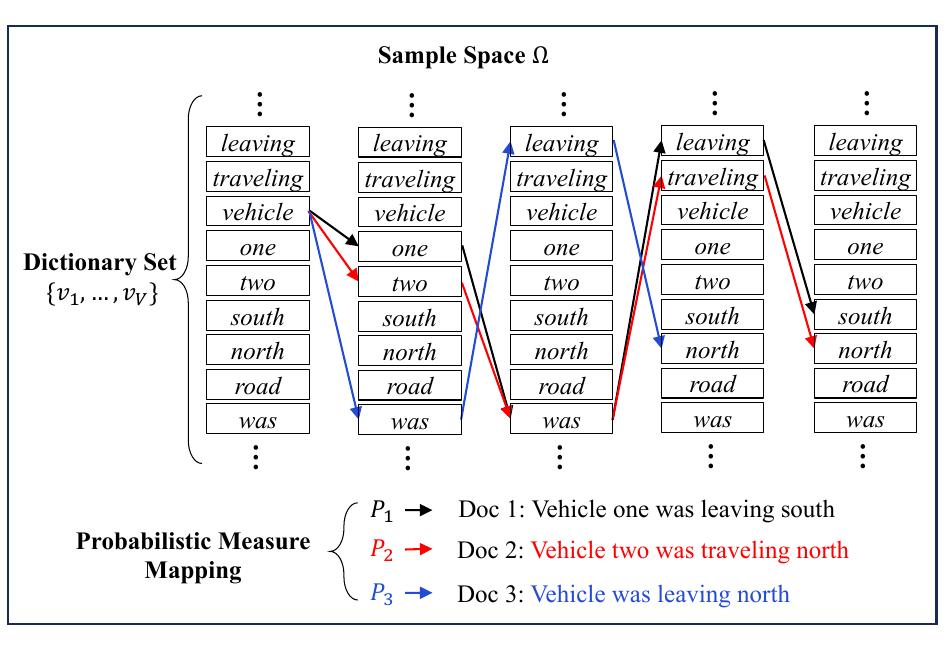}
\caption{The random variable $D$ generated from a probability space $(\Omega, \mathcal{F}, \mathrm{P})$ based on the dictionary set $\{v_1,...v_V\}$ under different probabilistic measure mapping $\mathrm{P}$}
\end{figure}

In {the context of this study, we assume that the} two collections of documents, group $A$ and $B$, share the same dictionary set but differ due to distinct probabilistic measures mapping $\mathrm{P}_A$ and $\mathrm{P}_B$. These differing measures result in distinct probability distributions across the two groups. The construction of probability measures for a document $D$, composed of words $\{W_1,...,W_T\}$, follows the principles of the $n$-gram model \citep{bengio2000neural}, where each word $W_t$ depends only on the $n-1$ preceding words. Accordingly, the probability of $D$, denoted as $\Pr(D)$, is expressed as the product of the probabilities of the first $n$ words and the conditional probabilities of the subsequent words given their preceding context. Thus, the likelihood of a document $D$ is decomposed following an auto-regressive model \citep{yang2019xlnet} as:
\begin{equation}
\label{pro_document}
\begin{aligned}
\Pr(D) =   \Pr(W_1,...W_n) \times \prod_{t=n+1}^{T} \Pr (W_t| H_t). 
\end{aligned}
\end{equation}
{where $ H_t = \{W_{\pi(1)},...,W_{\pi(n)}\} \setminus \{W_t\}$.}

\begin{figure*}[!ht]
\centering
\includegraphics[scale=0.7]{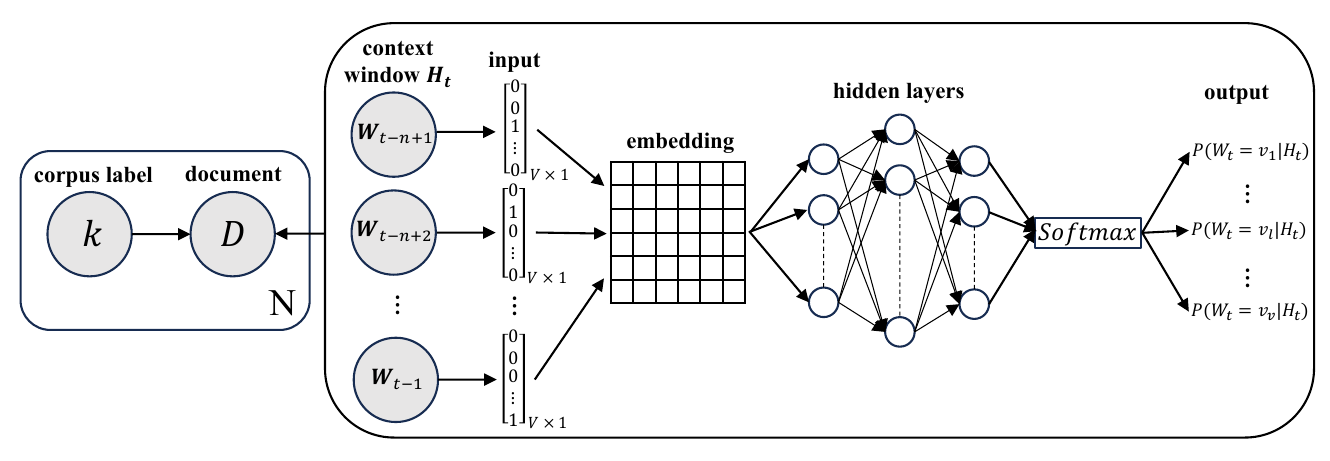}
\caption{The autoregressive neural network language model.\label{fig: diagram NNLM}}
\end{figure*}

An \textit{exchangeability} condition is assumed for the initial $n$ words, meaning that any random permutation of the sequence $(W_1,...,W_n)$ does not alter its semantic meaning. This condition can be expressed as the probability statement $\Pr(W_1,...,W_n) = \Pr(W_{\pi(1)},...,W_{\pi(n)})$ for any permutation $\pi$ of the set $\{1, ..., n\}$. The \textit{exchangeability} assumption suggests that the specific order of these initial $n$ words does not affect the document's initial semantic meaning. A relevant example is the use of the first $n$ words as a prompt for a Large Language Model (LLM). This assumption holds as LLMs can interpret and respond accurately to prompts, regardless of the sequence in which words or phrases are presented. Under this assumption, the joint log-likelihood of the document’s initial sequence can be decomposed as:
\begin{equation}
\begin{aligned}
\label{intiation}
    \Pr(W_1,...W_n) &= \prod_{t=1}^n \Pr \left(W_t\mid H_t\right), 
 \end{aligned}
\end{equation}

Given $H_t$, the order of the word will not affect the semantic meaning. The log-likelihood of any individual document can be expressed as a product of iterations over the conditional probability: 
\begin{equation}
\label{SampleDocumentLikelihood}
\begin{aligned}
       \log \Pr (D)  = \sum_{t=1}^{T}   \log \Pr (W_t |H_t ), 
\end{aligned}
\end{equation}
where the $ H_t = \{W_{\pi(1)},...,W_{\pi(n)}\} \setminus \{W_{t}\}$ if  $n\geq t\geq 1$. If the position $t$ satisfies   $ n+1 \leq t \leq T$,  the context is given by $H_t=   \{W_{t-n+1},..., W_{t-1}\} $.


\section{Two-Sample Text Test}
\label{sec: two sample text test}
\subsection{Estimation of the entropy for a document}
\label{subsec: test statistic}
This section {presents} the quantitative metrics to characterize documents and compare differences between two sets of documents. In information theory, entropy serves as a fundamental metric to quantify the average level of information contained within random variables. Entropy has been widely used to measure information levels in computational linguistics \citep{kontoyiannis1998nonparametric,moradi1998entropy} and evaluate the performance of large language models \citep{radford2019language,brown2020language}. This study extended the aforementioned works \citep{moradi1998entropy,radford2019language} to quantify text dissimilarity using entropy at the document level.

Entropy provides insight into the underlying generation process by capturing the probabilistic mapping of conditional generators. To measure the expected entropy of a document for a given corpus, denoted as \(\mu\), we define it as follows: 
\begin{equation}
 \mu = \underset{D\sim\Omega}{\mathbb{E}}\left[-\log \Pr(D)\right] = {\mathbb{E}} \Big[\sum_{t=1}^{T}   -\log \Pr (W_t |H_t )   \Big], 
\end{equation}
where $\mu$ serves as a key metric for analyzing the generative process through the \textit{logical flow}, with $H_t, W_t\sim \Omega$.
 
Given two collections of documents, the two-sample text test is applied to determine whether the entropy is equivalent for sources A and B. The null hypothesis and the alternative hypothesis are formulated as follows: 
\begin{equation}
\begin{aligned}
&H_0: \mu_A = \mu_B \\ 
&H_N: \mu_A \neq \mu_B
\end{aligned}
\end{equation}

Let $d_{k,j}$ denote the $j$th document from collection $k$ where $k=A $ or $k=B$. Denote $w_{k,j,t}$ as the $t$th word in the $j$th document of group $k$, and $T_{k,j}$ be the length of document $d_{k,j}$. Each document consists of a sequence of words, expressed as $d_{k,j} = \{w_{k,j,1}, \cdots ,w_{k,j,T_{k,j}}\}$. The document collections for groups A and B are denoted as ${d_{A,1}, \ldots, d_{A,N_A}}$ and ${d_{B,1}, \ldots, d_{B,N_B}}$, respectively, where $N_A$ and $N_B$ represent the total number of documents in the collections $A$ and $B$. Given the observed samples of documents $d_{k,1},...,d_{k,N_k}$, the population parameter \(\mu\) can be estimated at the sample level  through the following: 
\begin{equation}
\label{SampleLevelEntropy}
\widehat{\mu}_k = - \frac{1}{N_k}\sum_{j=1}^{N_k}\log \widehat{\Pr}(d_{k,j}).   \\
\end{equation}
 
The effect size $\mu_A-\mu_B$ is calculated by the difference between the empirical estimates $\widehat{\mu}_A-\widehat{\mu}_B$. The sample level estimates $\widehat{\mu}_A$ and $\widehat{\mu}_B$ are calculated using Equation~\ref{SampleLevelEntropy}. A larger observed difference between $\widehat{\mu}_A$ and $\widehat{\mu}_B$ suggests a higher likelihood of significant differences between document collections $A$ and $B$. Test statistics are constructed by normalizing the effect size $\widehat{\mu}_A-\widehat{\mu}_B$ against the variances associated with each group. To compute the denominator, the variance of the log-likelihood for each document at the sample level is calculated as follows:  
\begin{equation} \label{SampleVarianceDelta}
\widehat{\sigma}^2_{k} = \frac{1}{N_{k}}\sum_{j=1}^{N_{k}}\Bigg[  \frac{1}{N_k}\sum_{j=1}^{N_k}\log \widehat{\Pr}(d_{k,j}) - \log \widehat{\Pr}(d_{k,j})\Bigg]^2
\end{equation}
The test statistic $\Lambda_{A,B}$ is constructed as follows: 
\begin{eqnarray}
    \Lambda_{A,B} = \frac{\widehat{\mu}_A-\widehat{\mu}_B}{\sqrt{\widehat{\sigma}_{A}^2/N_A + \widehat{\sigma}_{B}^2/N_B }} 
\end{eqnarray}

The calculation of $\Lambda_{A,B}$ depends on the approximation of $\widehat{\Pr}(d_{k,j})$. However, estimating $\widehat{\Pr}(d_{k,j})$ using word frequency-based methods for discrete data encounters the \textit{curse of dimensionality}. The computational cost will be inflated as $\mathcal{O}(V^n)$, where $V$ is the vocabulary size and $n$ is the context window size. As $V$ increases significantly, often in the hundreds, the resulting computational complexity can quickly exceed practical computational limits. 

The Neural Network-based Language Model (NNLM) addresses this issue by transforming discrete word data into continuous embeddings \citep{bengio2000neural}. This approach not only makes the computation of conditional probabilities feasible but also captures the semantic meaning of words within their context \citep{brown2020language}. To estimate $\log\Pr(d_{k,j})$ using NNLM, each word $w_{k,j,t}$ is first represented as a one-hot encoded vector $\boldsymbol{w}_{k,j,t}$, where a single element is 1 and all others are 0. This vector is then transformed into a continuous embedding $\mathcal{E}(\boldsymbol{w}_{k,j,t})$. The function $\mathcal{E}$ maps the original $V \times 1$ vector to a much smaller $q \times 1$ vector, where $q<V$. The embeddings of the context words, $\mathcal{E}(\boldsymbol{w}_{k,j,t-n+1}), \ldots, \mathcal{E}(\boldsymbol{w}_{k,j,t-1})$ are stacked to form a context vector of size $(n-1)q \times 1$.

Figure~\ref{fig: diagram NNLM} illustrates the autoregressive structure of the neural network-based language model for textual data. The input document is first tokenized, breaking it down into individual words. Each word is then mapped to a unique vector representation through an embedding matrix, which is learned during the training process. This matrix transforms each word into a dense, fixed-dimension representations. The hidden layers apply a fully connected neural layer with a nonparametric mapping $f: \mathbb{R}^{(n-1)q\times1} \rightarrow \mathbb{R}^{V\times 1}$, followed by a Softmax layer. Given the context $h_{k,j,t}$, the probability of the next word is approximated using the function: $\operatorname{Softmax}\left\{f\big[\mathcal{E}(\boldsymbol{w}_{k,j,t-n+1}),...,\mathcal{E}(\boldsymbol{w}_{k,j,t-1})\big]\right\}$. For simplicity, we define $g(\cdot)$ as a composite function that integrates the mapping function $f$ with the embedding function $\mathcal{E}$, as follows: 
\begin{equation}
\begin{aligned}
    &g(h_{k,j,t}) = g(\boldsymbol{w}_{k,j,t-n+1},...,\boldsymbol{w}_{k,j,t-1}) \\
    &= \operatorname{Softmax}\Big\{f\big[\mathcal{E}(\boldsymbol{w}_{k,j,t-n+1}),...,\mathcal{E}(\boldsymbol{w}_{k,j,t-1})\big]\Big\} . 
\end{aligned}
\end{equation}
The output of $g(h_{k,j,t})$ is a $V\times1$ vector, enabling the approximation of the log-probability $\log \Pr (W_t = w_{k,j,t}|H_t = h_{k,j,t})$ as given by the following equation:
\begin{equation}
\begin{aligned}
\label{ImportantIntermediate}
    \log & \Pr (W_t = w_{k,j,t}|H_t = h_{k,j,t})  \\
    & =(\boldsymbol{w}_{k,j,t})^{\top} \log\big[g(h_{k,j,t})\big],
\end{aligned}
\end{equation}
Using Equation~\ref{ImportantIntermediate} and \ref{SampleDocumentLikelihood}, we normalize the log-likelihood by the document length to reduce sensitivity to varying lengths, as follows:
\begin{equation}
\label{IndividualLogLikelihood}
     \log \Pr(d_{k,j}) =  \frac{1}{T_{k,j}}\sum_{t=1}^{T_{k,j}}(\boldsymbol{w}_{k,j,t})^{\top}\log\big[g(h_{k,j,t})\big],
\end{equation}
Using Equation~\ref{IndividualLogLikelihood} and ~\ref{SampleLevelEntropy}, along with the empirical function $\widehat{g}(.)$, the estimates $\widehat{\mu}_A $ and $\widehat{\mu}_B$ are calculated as follows:
\begin{equation}
\label{SampleMu-Use}
\begin{aligned}
\widehat{\mu}_k = - \frac{1}{N_{k}}\sum_{j=1}^{N_{k}}\Big\{\frac{1}{T_{k,j}}\sum_{t=1}^{T_{k,j}}(\boldsymbol{w}_{k,j,t})^{\top}\log\big[\widehat{g}(h_{k,j,t})\big]\Big\}.     
\end{aligned}
\end{equation}

The empirical estimation of the variance can be obtained by combining Equation~\ref{SampleVarianceDelta} and Equation~\ref{IndividualLogLikelihood}, as follows: 
\begin{equation} \label{SampleVariance-Use}
\widehat{\sigma}^2_{k}  = \frac{1}{N_{k}}\sum_{j=1}^{N_{k}}\Big\{-\frac{1}{T_{k,j}}\sum_{t=1}^{T_{k,j}}(\boldsymbol{w}_{k,j,t})^{\top}\log\big[\widehat{g}(h_{k,j,t})\big]-\widehat{\mu}_k\Big\}^2, 
\end{equation}
The language model $g(\cdot)$ is trained by minimizing the following objective function:
\begin{equation}
    g^*(\cdot) = \underset{g}{\operatorname{\text{argmin}}} \ \mathbb{E}\Big[-\sum_{t=1}^{T_k}\boldsymbol{W}_t^{\top}\log\big(g( H_t)\big)\Big], 
\end{equation}
where $\boldsymbol{W}_t$ is the one-hot encoding vector for word $W_t$, and $H_t$ represents the contextual variable for $W_t$. $T_k$ denotes the expected document length for group $k$. We assume that for any collection of documents $k$, condition $g^*_k(H_t) = g^*(H_t) \quad \forall H_t \sim k$ holds, ensuring that conditional probability estimates remain consistent across different information sources. This invariance property has been shown in machine learning applications \citep{brown2020language}. The optimization objective is to identify a unique and invariant $g^*_k(\cdot)$ that satisfies the following condition, where $H_t$ is from the group $k$:
\begin{equation}
\begin{aligned}
    g^*_k(\cdot) = \underset{g}{\operatorname{\text{argmin}}} \ \mathbb{E}\Big\{-\sum_{t=1}^{T_k}\boldsymbol{W}_t^{\top}\log\big[g( H_t)\big]\Big\}
\end{aligned}
\end{equation}
The empirical function $\widehat{g}(\cdot)$, integrates an embedding layer $\mathcal{E}$, a fully connected layer $f_{\mathcal{M}}$ parameterized by $\mathcal{M}$, an activation function $a(\cdot)$ (typically set as a ReLU function), and a Softmax layer. This function is optimized through:
\begin{equation}
\begin{aligned}
\widehat{g}(\cdot) & = \underset{\mathcal{C},\mathcal{M}}{\operatorname{argmin}}\sum_{k\in[A,B]}\sum_{j=1}^{N_{k}}\sum_{t=1}^{T_{k,j}}(\boldsymbol{w}_{k,j,t})^{\top} \\ 
& \times -\log\left[\operatorname{Softmax}\left\{f_{\mathcal{M}}\big[\mathcal{E}(h_{k,j,t})\big]\right\}\right], 
\end{aligned}
\end{equation}
where $f_{\mathcal{M}}(\boldsymbol{x}) = \boldsymbol{M}_L a(\boldsymbol{M}_{L-1} \cdots a(\boldsymbol{M}_1 \boldsymbol{x}))$ represents a neural network with $L$ layers and activation function $a(\cdot)$. The parameters of each fully connected layer are denoted by $\boldsymbol{M}_{1},\cdots \boldsymbol{M}_{L} \in \mathcal{M}$. The embedding function $\mathcal{E}(h_{k,j,t})$ maps the context $h_{k,j,t}$ into a vector space, with $ \mathcal{E}(h_{k,j,t})\in \mathbb{R}^{(n-1)d\times1} $. The final output, $\operatorname{Softmax}\left\{f_{\mathcal{M}}\big[\mathcal{E}(h_{k,j,t})\big]\right\} \in \mathbb{R}^V$, provides a probability distribution over the vocabulary in a dictionary set.

\subsection{Asymptotic behavior of the two-sample text test}
\label{subsec: support theory}
In this section, we examine the asymptotic behavior of the test statistics $\Lambda_{A,B}$. The behavior of $\Lambda_{A,B}$ is influenced by the differences $\widehat{\mu}_A - \mu_A$ and $\widehat{\mu}_B - \mu_B$.  We decompose $\widehat{\mu}_k-\mu_k$ into two components $U_{k,1}$ and $U_{k,2}$, as follows:
\begin{equation}
\begin{aligned}
\widehat{\mu}_k-\mu_k & =U_{k.1} + U_{k,2}, 
\end{aligned} 
\end{equation}
For simplicity, we denote the $U_{k,1}$ term as follows:
\begin{equation}
\small
\begin{aligned}
    U_{k,1} = &\frac{1}{n_k}\sum_{j=1}^{n_k}\Bigg\{-\frac{1}{T_{k,j}}\sum_{t=1}^{T_{k,j}}(\boldsymbol{w}_{k,j,t})^{\top}\log\big[\widehat{g}(h_{k,j,t})\big]\Bigg\}\\
    &-\mathbb{E}\Big\{-\frac{1}{T_K}\sum_{t=1}^{T_k}\boldsymbol{W}_t^{\top}\log\big[\widehat{g}( H_t)\big]\Big\}, 
\end{aligned}
\end{equation}
where $T_k$ denotes the expected length of documents in group $k$. The term $U_{k,2}$ is defined as follows, and for simplicity we define $\mathbb{E}\Big\{-\frac{1}{T_k}\sum_{t=1}^{T_k}\boldsymbol{W}_t^{\top}\log\big[\widehat{g}(H_t)\big]\Big\}$ as the expected entropy based on the empirical function $\widehat{g} (\cdot)$. 
\begin{equation}
\begin{aligned}
U_{k,2} = \mathbb{E}\Big\{&-\frac{1}{T_k}\sum_{t=1}^{T_K}\boldsymbol{W}_t^{\top}\log\big[\widehat{g}(H_t)\big]\Big\} \\
&- \underset{D\sim k}{\mathbb{E}}\big[-\log \Pr(D)\big]. 
\end{aligned}
\end{equation}
Following the notation in \citep{dasgupta2008asymptotic}, $x \lesssim y$ indicates that $x \leq cy$ for some constant $c$, and $a_n \asymp b_n$ denotes that $a_n$ and $b_n$ are of the same order. The notation $a_n=o\left(b_n\right)$ implies $
\underset{n \rightarrow \infty}{\lim}{\frac{a_n}{b_n}=0}$, while $O_p(.)$ denotes stochastic boundedness.

For the $U_{k,1}$ part, we show that $\frac{\sqrt{n_k}}{\widehat{\sigma}_{k}} U_{k,1}$ converges to a standard normal distribution as $N_{k}$ becomes sufficiently large. The asymptotic behavior of $U_{k,1}$ is detailed in Theorem~\ref{ThmT11}. For group $k$, the dataset is divided into an inference set and an estimation set. The inference set has a size of $n_k$, while the estimation set size is given by ${n_k}^{1/\gamma}$, where $0< \gamma <1$, ensuring that $n_k + {n_k}^{1/\gamma} = N_k$. A detailed discussion on the data-splitting for statistical inference is provided in Section~\ref{subsec: estimation and inference}.

{\thm  {\label{ThmT11}}{Assuming that there $\exists$ $\delta>0$, such that, as $n_k$ goes sufficiently large, for any $j$, the condition $\left(\frac{1}{n_{k}}\right)^\delta \mathbb{E}(\Delta_{k,j})^{2+2 \delta} \rightarrow 0$, 
where $\Delta_{k,j}= -\frac{1}{T_{k,j}}\sum_{t=1}^{T_{k,j}}(\boldsymbol{w}_{k,j,t})^{\top}\log\big[\widehat{g}(h_{k,j,t})\big]$. With the estimation and inference sets, the following result holds:
\begin{equation}
\frac{\sqrt{n_k}}{\widehat{\sigma}_{k}} U_{k,1} \xrightarrow{d} \mathcal{N}(0,1)    
\end{equation}
where $\mathcal{N}\left(0, 1\right)$ denotes the standard normal distribution.}} 

For the $U_{k,2}$ part, the asymptotic behavior depends on how well $\widehat{g}($.$)$ approximates the function $g($.$)$. \citep{bos2022convergence} provides an in-depth discussion on the approximation error of $\widehat{g}($.$)$ for multi-class output using a Softmax activation function. Following the notation in \citep{bos2022convergence}, let $L$ denote the depth of the neural network, $\beta$ the Holder smoothness index, and $q$ the input dimension of $\widehat{g}(\cdot)$. 

We define $\Pr(W_t = v_l | H_t = h_{k,j,t})$ as $P_{v_l}(h_{k,j,t})$. A challenge arises when conditional probabilities diminish, driving the log-likelihood $\log\Pr(W_t = v_l | H_t = h_{k,j,t})$ toward negative infinity. To address this, a truncation parameter $B$ is introduced, ensuring that the estimated conditional probabilities are bounded within the range $ P_{v_l}(h_{k,j,t}) \in [e^{-B}, 1]$ for any $v_l \in \{v_1,...,v_V\}$ and $B \geq 2$. 

Additionally, a parameter $\alpha$ controls ``the size of the set" on which the conditional probabilities are small, with the constraint $\operatorname{Pr}\left \{P_{v_k}(h_{k,j,t})\leq \epsilon \right\} \leqslant C \epsilon^{\alpha}$, where $C$ is some constant \citep{bos2022convergence}. The expected document length for group $k$ is denoted as $T_k$, and the sample size used for approximating the empirical function is on the order of $S_k$. Supplementary material detailed the regularity condition for the neural network, the related technical lemma, and the proof of Theorem~\ref{Thm12}. 

{\thm \label{Thm12} Given $T_k = o(n_k^{\tau}) $ and $\tau< \frac{(1+\alpha)\beta}{d}$, such that $T_k \lesssim n_k^{\frac{(1+\alpha)\beta}{d}}$, where $q$ is the input dimension and $0<\beta\leq1$ represents the Holder smoothness index. In data-splitting procedure, the parameter $\gamma$ controls the size of the inference and estimation sets, satisfying $\gamma<\frac{2 [(1+\alpha)\beta-q\tau]}{(1+\alpha)\b geta+q}$. Under these conditions, the following holds: 
 \begin{equation}
 \text{As} \quad N_k\rightarrow \infty\quad 
      \text{,} \quad \frac{\sqrt{n_k}}{\widehat{\sigma}_{k}}  U_{k,2}\rightarrow0
 \end{equation}
 where $n_k$ denotes the size of the inference set, and the relationship $n_k + {n_k}^{1/\gamma} = N_k$  holds.}
 
{\thm  \label{maintheorem}
The test statistic $\Lambda_{AB}$ evaluates the difference between $\widehat{\mu}_A$ and $\widehat{\mu}_B$. Under the null hypothesis that $\mu_A = \mu_B$, the test statistics is asymptotically distributed as:
\begin{equation}
\frac{\widehat{\mu}_A-\widehat{\mu}_B}{\sqrt{{\widehat{\sigma}_{A}^2}/{n_{A}} + {\widehat{\sigma}_{B}^2}/{n_{B}}}}  \xrightarrow{d} \mathcal{N}\left(0, 1\right)
\end{equation}
where $\mathcal{N}\left(0, 1\right)$ denotes a standard normal distribution.  }

Theorem~\ref{Thm12} examines the convergence properties of the test. The parameter $q$ controls the dimensionality of the input variables, with larger values of $q$ resulting in slower convergence rates. As discussed in \citep{bos2022convergence}, $q$ can represent the dimensionality of different composite structures; in the context of language models, it often corresponds to embedding dimension. Based on Theorem ~\ref{ThmT11} and \ref{Thm12}, Theorem~\ref{maintheorem} establishes that  $\Lambda_{AB}$ asymptotically converge to a standard normal distribution. This result provides a theoretical basis for the hypothesis test, enabling statistical inference on textual data to be as straightforward as a two-sample t-test. It eliminates the need for computationally intensive nonparametric methods, such as permutation tests commonly used in MMD calculations \citep{jitkrittum2016interpretable}.

\subsection{Data-splitting for statistical inference}
\label{subsec: estimation and inference}
Given the higher model complexity of deep neural networks, the approximation of the function $\widehat{g}(\cdot)$ is prone to overfitting, which can introduce bias in the estimation of $\widehat{\mu}_k$. To mitigate the risk of overfitting, we employ a data-splitting approach. Each collection of documents is divided into two distinct sets: the \textit{Estimation} set, $k_{\text{Est}} = \{d^{\text{Est}}_{k,1},...,d^{ \text{Est}}_{k,N_{k,\text{Est}}}\}$, used for model training, and the \textit{Inference} set, $k_{\text{Inf}} = \{d^{\text{Inf}}_{k,1},...,d^{ \text{Inf}}_{k,N_{k,\text{Inf}}}\}$, used for test inference. This approach has proven effective in various research on nonparametric statistical inference \citep{liu2022multiple, liu2022projection, dai2022significance}. 

Algorithm~\ref{Algorithm} provides the estimation and inference procedure for constructing test statistics. 
The \textit{Estimation} and \textit{Inference} sets are designed to be mutually exclusive, ensuring that $k_{\text{Est}} \cup k_{\text{Inf}} =k$ and $k_{\text{Est}} \cap k_{\text{Inf}} = \emptyset$. The data set is first partitioned into the \textit{Estimation} set, which is used for model training, for example, to learn the weights of a neural network. Once model training is complete, the algorithm proceeds to the \textit{Inference} phase, where the test results are calculated. In the case of two samples, where $k \in \{A, B\}$, the estimation sets $A_{\text{Est}}$ and $B_{\text{Est}}$ are used to fit the function $\widehat{g}(\cdot)$. The fitted function is then applied to the inference sets $A_{\text{Inf}}$ and $B_{\text{Inf}}$ to estimate the difference $\widehat{\mu}_A - \widehat{\mu}_B$. According to Theorem~\ref{PowerLoss}, when differences $\delta = \mu_A -\mu_B$ or sample size becomes sufficiently large, the asymptotic power of the test approaches 1. 

{\thm \label{PowerLoss} Let $\delta =\mu_A - \mu_B $ and $\sigma = \sqrt{{\sigma_A^2}/{n_{A}} + {\sigma_B^2}/{n_{B}}}$. The power of the test, denoted as $\beta(\delta)$, is given by: 
\begin{equation}
\begin{aligned}
& \lim _{n_* \rightarrow \infty} \sup \beta(\delta)=\Phi\left(-z_\alpha+{\delta}/{\sigma}\right) \quad \\ 
& \text{and} \quad \lim _{\delta \rightarrow \infty} \lim _{n_* \rightarrow \infty} \sup \beta(\delta)=1        
\end{aligned}
\end{equation}
where $ n_* = \min\{n_A,n_B\}$, and $z_\alpha=\Phi^{-1}(1-\alpha)$ is the critical value from the standard normal distribution at significance level $\alpha$. }

\begin{algorithm}
\caption{\label{Algorithm}Estimation and Inference for Text Test}
\begin{algorithmic}
\STATE {\bf Step 1:}  Randomly split  group A into $A_{\text{Est}}$ and $ A_{\text{Inf}}$, group B into  $B_{\text{Est}}$ and $ B_{\text{Inf}}$.
\STATE {\bf Step 2:}   Optimize the NNLM using the \textit{Estimation} sets $A_{\text{Est}}$ and $B_{\text{Est}}$ by minimizing the following objective:
\begin{equation} 
\begin{aligned}
\widehat{g}(\cdot) &= \underset{\mathcal{C},\mathcal{M}}{\operatorname{argmin}}\sum_{k\in[A,B]}\sum_{j=1}^{N_{k,
\text{Est}}}\sum_{t=1}^{T_{k,j}}(\boldsymbol{w}_{k,j,t})^{\top} \\ & \times -\log\left[\operatorname{Softmax}\Big\{f_{\mathcal{M}}\big[\mathcal{E}(h_{k,j,t})\big]\Big\}\right]
\end{aligned}
\end{equation}
\hspace{0.5em}
\STATE {\bf Step 3:}   Use the \textit{Inference} sets $A_{\text{Inf}}$ and $B_{\text{Inf}}$ to compute $\widehat{\mu}_A$ and  $\widehat{\mu}_B$ using Equation~\ref{SampleMu-Use} and estimate the variances using Equation~\ref{SampleVariance-Use}. 

\hspace{0.5em}
 \STATE {\bf Step 4:}  Compute the test statistic: 
 \begin{equation}
     \Lambda_{A,B} = \frac{\widehat{\mu}_A-\widehat{\mu}_B}{\sqrt{\widehat{\sigma}_{A}^2/N_{A,\text{Inf}} + \widehat{\sigma}_{B}^2/N_{B,\text{Inf}} }} 
 \end{equation}

 \STATE {\bf Step 5:}    Calculate the \textit{p}-value as $ \text{\textit{p}-value} = 1- \Phi(|\Lambda_{A,B} |)$ , where  $ \Phi (x)={\frac {1}{\sqrt {2\pi }}}\int _{-\infty }^{x}e^{-t^{2}/2}\operatorname{d}t$. Reject the null hypothesis if $\text{\textit{p}-value} \leq \alpha$. 
\hspace{0.5em}
\label{mainallgorithm}
\end{algorithmic}
\end{algorithm}

\subsection{Theorem on power loss}
\label{subsec: power enhancement}
Theorem~\ref{PowerLoss} highlights that the data-splitting procedure can results in a loss of statistical power, as only a subset of the dataset is used for inference.  This issue becomes particularly challenging when the sample sizes $N_A$ and $N_B$ are limited. To mitigate this, a multiple data-splitting strategy is employed, as supported by various empirical studies \citep{liu2022multiple,dai2022significance,liu2019cauchy}. This approach involves repeating the data-splitting procedure $M$ times, generating  corresponding \textit{p}-values as $\text{p}_1,...\text{p}_M$ in each iteration. By aggregating these \textit{p}-values into a unified decision, the strategy maximizes the utilization of available data, thereby enhancing the statistical power of the hypothesis test. This study adopts the following \textit{p}-value combination methods to improve the power of the two-sample text test. 

The first approach is the Cauchy combination method \citep{liu2019cauchy}, which  leverages the Cauchy distribution's robustness to dependencies among the \textit{p}-values. According to \citep{liu2019cauchy}, the null hypothesis is rejected based on the multiple splits $\text{p}_1,...\text{p}_M$ if 
$
    \sum_{i=1}^M \tan[(0.5-\text{p}_i)\pi] \geq Mc_{\alpha}
$
where $c_{\alpha}$ is the upper $\alpha$-quantile of the standard Cauchy distribution. 

The second combination approach is the Multiple-splitting Projection Test (MPT) proposed by \citep{liu2022multiple}. The MPT provides a framework for hypothesis testing that accounts for dependencies among test statistics. Let $Z_i = \Phi^{-1}(\text{p}_i)$ for $i=1,...,M$,  where  $ \Phi (x)={\frac {1}{\sqrt {2\pi }}}\int _{-\infty }^{x}e^{-t^{2}/2}\operatorname{d}t$ is the cumulative distribution function of the standard normal distribution. Based on the \textit{exchangeability} condition for $\text{p}_1,...,\text{p}_M$, \citep{liu2022multiple}   assume a consistent dependency structure among   $Z_1,...,Z_M$, with $\operatorname{Cov}(Z_i,Z_j) = \rho$. The null hypothesis is rejected if
$
        \frac{ \frac{1}{m}\sum_{i=1}^M \Phi^{-1}(p_i)}{\sqrt{(1+(M-1)\widehat{\rho})/M}} \geq c(M,\alpha/2)
$
where $c(M,\alpha/2)$ is a critical threshold depending on $M$ and the type I error $\alpha$.  

In \citep{liu2022multiple}, there are two different methods for estimating the parameter $\widehat{\rho}$ to determine threshold for rejecting the null hypothesis. The first method approximates $\widehat{\rho}$ as $\widehat{\rho} = \max\{0,1-S_Z^2\}$, where $S_Z^2$ is the sample variance of $Z_1,...,Z_M$. We refer to this approach as ``MPT1" in the following analysis. The second method, termed ``MPT2", approximates $\widehat{\rho}$ as $\widehat{\rho}=\max \left\{0,1-(M-1) S_Z^2 / \chi_{M-1}^2(1-\beta)\right\}$, where $\chi_{M-1}^2(1-\beta)$ is the upper $(1-\beta)$ quantile of a chi-square distribution with $M-1$ degrees of freedom.  

\section{Numerical Results}
\label{sec: numerical results}
We first conducted Monte Carlo simulations in various data generation processes to assess the effectiveness of our proposed methodology. The neural network models were trained using PyTorch version 2.5.1. We compare the Maximum Mean Discrepancy (MMD) kernel-based test \citep{li2019optimality, jitkrittum2016interpretable}, where documents are represented using a vector space model \citep{wang2021sequential}. In our study, the MMD test was configured according to the hyperparameter settings suggested in \citep{li2019optimality}. 

The second comparative method combines the Latent Dirichlet Allocation (LDA) model \citep{blei2003latent} with the Hotelling $T^2$ test \citep{hotelling1992generalization}. This benchmark approach utilizes LDA's topic modeling capabilities by analyzing each document for its probability of association with predefined topics. The resulting topic assignment probabilities are then used as input for the Hotelling $T^2$ test to perform two-sample comparisons. Configuring the LDA model requires specifying the number of topics in advance. In our simulations, this approach is referred to ``LDA", with the number of topics set to 10. 

The third comparative method is the Classification Accuracy Two-sample Test (C2ST) proposed by \citep{kim2021classification}. C2ST splits the data into two parts: one for training a classifier and the other for evaluating its performance. The underlying principle is that if the classifier achieves accuracy significantly above random chance, the null hypothesis is rejected. In our implementation, we use the random forest as the classifier and represent each document using the \textit{Word2Vec} approach. 

Performance evaluation metrics are \textit{power} and \textit{size}. \textit{Power} is the probability that the test correctly rejects the null hypothesis when the alternative hypothesis is true. Ranges from 0 to 1, with a higher power indicating a more effective test. \textit{Size} refers to the probability of a Type I error, representing the probability of incorrectly rejecting the null hypothesis when it is true. Ideally, the size should be controlled at approximately 0.05 when the significance level is set to $\alpha=0.05$. 

\subsection{Simulation studies}
\label{subsec: simulation results}
Textual data is modeled as a sequence of discrete random variables. Following recent studies \citep{chen2022inferential}, we simulate the nature of text data through a latent generation process, first generating latent variables and then transforming them into a series of categorical values. For each simulation round, Algorithm~\ref{Algorithm} is repeated $M$ times, utilizing the \textit{p}-value combination approaches outlined in Section~\ref{subsec: power enhancement} to increase statistical power. This procedure requires training a neural network for each split. We set $M=10$ in the numerical studies and repeated each of the simulations 1,000 times. 

\begin{table*}[h]
\footnotesize
\caption{\label{SIMU1Power}Power comparison under AR covariance matrix settings}
\centering
\small
\begin{tabular}{c|c|c|cccc|ccc}
\hline
\hline
Param  S                  & Param V              & Param $\delta$ & Cauchy & MPT1  & MPT2  & Single & LDA   & C2ST  & MMD   \\
\hline
\multirow{8}{*}{4} & \multirow{4}{*}{25}     & 0.1    & 0.378  & 0.281 & 0.297 & 0.265  & 0.057 & 0.089 & 0.050  \\
                   &      & 0.2    & 0.612  & 0.537 & 0.553 & 0.489  & 0.236 & 0.185 & 0.141 \\
                   &      & 0.3    & 0.753  & 0.705 & 0.716 & 0.649  & 0.453 & 0.310  & 0.296 \\
                   &      & 0.4    & 0.847  & 0.817 & 0.821 & 0.768  & 0.665 & 0.418 & 0.548 \\ \cline{2-10}
                   & \multirow{4}{*}{50}      & 0.1    & 0.363  & 0.275 & 0.295 & 0.255  & 0.050  & 0.106 & 0.061 \\
                   &       & 0.2    & 0.572  & 0.490  & 0.512 & 0.471  & 0.141 & 0.180  & 0.128 \\
                   &      & 0.3    & 0.715  & 0.672 & 0.688 & 0.637  & 0.327 & 0.290  & 0.310  \\
                   &      & 0.4    & 0.844  & 0.791 & 0.799 & 0.759  & 0.564 & 0.380  & 0.521 \\ 
\hline 
\multirow{8}{*}{3} & \multirow{4}{*}{25}     & 0.1    & 0.307  & 0.225 & 0.244 & 0.199  & 0.057 & 0.084 & 0.062 \\
                   &      & 0.2    & 0.520   & 0.443 & 0.452 & 0.398  & 0.179 & 0.160  & 0.113 \\
                   &      & 0.3    & 0.655  & 0.599 & 0.602 & 0.539  & 0.412 & 0.229 & 0.262 \\
                   &      & 0.4    & 0.791  & 0.747 & 0.757 & 0.698  & 0.644 & 0.354 & 0.485 \\
                   \cline{2-10}
                   & \multirow{4}{*}{50}     & 0.1    & 0.250   & 0.184 & 0.199 & 0.179  & 0.044 & 0.069 & 0.062 \\
                   &      & 0.2    & 0.501  & 0.426 & 0.439 & 0.388  & 0.114 & 0.156 & 0.107 \\
                   &      & 0.3    & 0.640   & 0.570  & 0.583 & 0.518  & 0.309 & 0.219 & 0.240  \\
                   &      & 0.4    & 0.759  & 0.715 & 0.732 & 0.670   & 0.500   & 0.338 & 0.467 \\
\hline 
\multirow{8}{*}{2} & \multirow{4}{*}{25}     & 0.1    & 0.193  & 0.129 & 0.134 & 0.127  & 0.055 & 0.082 & 0.056 \\
                   &      & 0.2    & 0.371  & 0.300   & 0.314 & 0.288  & 0.117 & 0.077 & 0.116 \\
                   &      & 0.3    & 0.532  & 0.459 & 0.468 & 0.397  & 0.299 & 0.153 & 0.218 \\
                   &      & 0.4    & 0.616  & 0.559 & 0.562 & 0.522  & 0.500   & 0.233 & 0.409 \\ \cline{2-10}
                   &  \multirow{4}{*}{50}     & 0.1    & 0.183  & 0.123 & 0.129 & 0.128  & 0.057 & 0.062 & 0.064 \\
                   &      & 0.2    & 0.349  & 0.287 & 0.292 & 0.251  & 0.071 & 0.080  & 0.104 \\
                   &      & 0.3    & 0.494  & 0.421 & 0.437 & 0.367  & 0.175 & 0.145 & 0.247 \\
                   &      & 0.4    & 0.597  & 0.540  & 0.553 & 0.505  & 0.377 & 0.219 & 0.426 \\
                   \hline
                   \hline
\end{tabular}
\end{table*}

Each simulated document $D_i$ consists of $T_i$ words, represented as $D_i = \{w_{i,1},...,w_{i,T_i}\}$. The term $w_{i,t}$ denotes the $t$-th word in document $i$, selected from a dictionary set $\{v_1,...,v_V\}$, where $1 \leq t \leq T_i$. The latent variable $x^*_{i,t}$ at the position $t$ controls the semantic meaning at that position within the document $i$. The dependence between two words $w_{i,t}$ and $w_{i,t^{\prime}}$, is modeled by the dependence between their corresponding latent variables,  $x^*_{i,t}$ and $x^*_{i,t^{\prime}}$. 

Each word $w_{i,t}$ is drawn from a multinomial distribution with a single trial, defined as: 
\begin{equation}
\label{WordGeneration-Simulation}
    w_{i,t}\sim \text{MN}(p_{t,1},...,p_{t,V})
\end{equation}
where $\text{MN}(p_{t,1},\ldots,p_{t,V})$ represents a multinomial distribution with probabilities $p_{t,1},\ldots,p_{t,V}$. The probability $p_{t,l}$ corresponds to the likelihood that word $v_l$ occurs at position $t$, for $1\leq l\leq V$. This probability $p_{t,l}$ is derived from the latent variable $x^*_{i,t}$ as follows: 
\begin{equation}
\label{ProbGeneration-Simulation}
p_{t,l} = \frac{\exp(\beta_lx^*_{i,t})}{\sum_{l=1}^V\exp(\beta_l x^*_{i,t})}     
\end{equation}
The coefficients $\beta_1, \ldots, \beta_V$ control the selection of words, with certain words occurring more frequently, typically referred to as high-frequency words. To replicate this phenomenon, fixed values are assigned to the coefficients corresponding to the first 20\% words in the dictionary. A parameter $S>1$ is used to adjust for the presence of high-frequency words in the simulation:  
$
    \beta_1, \ldots, \beta_{[0.2V]} = S
$.
The coefficients for the remaining 80\% of words are drawn from a uniform distribution: 
$
\beta_{[0.2V]+1}, \ldots, \beta_{V} \sim \text{Uniform}(0,1)
$. 
The latent variables $x^*_{i,1},...,x^*_{i,T}$ are drawn from a multivariate Gaussian distribution $\mathcal{N}(\mathbf{0}_{1 \times T},\Sigma_{T \times T})$. The covariance matrix $(\Sigma_{T \times T})$ governs the dependencies among the latent variables, with each element $\Sigma_{p,q}$ at the $p$-th row and $q$-th column ($1\leq p,q \leq T$) controlling the relationship between $x^*_{i,p}$ and $x^*_{i,q}$. We implemented two types of covariance matrices to explore different word dependency structures, as suggested in \citep{liu2022multiple}.

\noindent \textbf{Setup 1:} The covariance matrix is structured using an autoregressive (AR) approach, where each element is defined as $\Sigma_{p,q} = \theta^{|p-q|}$ with the parameter $\theta\in (0,1)$ set for the simulation. This AR structure captures the decreasing dependency between words as the distance increases. It enables a comparative analysis of statistical power by using different values of $\theta_A$ for group A and $\theta_B$ for group B. In the simulation, we set the sample size to $N=100$ and the document length to $T=25$. We explored the effect of varying dictionary size $V$ by setting $V=25$ and $V=50$. The parameters $\theta_A$ and $\theta_B$ were defined as $\theta_A = 0.5 + \delta/2$ and $\theta_B = 0.5 - \delta/2$, respectively. For power comparison, $\theta_A$ and $\theta_B$ were adjusted across $\delta \in \{ 0.1,0.2,0.3,0.4\}$ and $S \in \{ 2,3,4\}$. 

Table~\ref{SIMU1Power} highlights the effectiveness of the \textit{p}-value combination strategy in enhancing test power in various settings. Our proposed method outperforms competing approaches, particularly when the dictionary size $V$ is larger. To evaluate the size of the hypothesis test, we set $\theta_A=\theta_B = 0.5$ and vary $S \in \{2,3,4\}$. As shown in Table~\ref{SIMU12Size}, our approach controls the size of hypothesis tests under various simulation setups. 
\begin{table*}
\footnotesize
\caption{\label{SIMU12Size}Size comparison under AR covariance matrix settings}
\centering
\begin{tabular}{c|c|c|cccc|ccc}
\hline
\hline
Setup                 & Param V                   & Param S & Cauchy & MPT1  & MPT2  & Single & LDA   & C2ST  & MMD   \\
\hline 
\multirow{6}{*}{AR} & \multirow{3}{*}{25} & 2 & 0.065  & 0.036 & 0.041 & 0.055  & 0.037 & 0.058 & 0.051 \\
                    &                     & 3 & 0.058  & 0.036 & 0.041 & 0.042  & 0.039 & 0.058 & 0.038 \\
                    &                     & 4 & 0.048  & 0.033 & 0.032 & 0.051  & 0.032 & 0.053 & 0.044 \\
\cline{2-10}
                    & \multirow{3}{*}{50} & 2 & 0.064  & 0.032 & 0.041 & 0.044  & 0.03  & 0.067 & 0.047 \\
                    &                     & 3 & 0.056  & 0.026 & 0.028 & 0.051  & 0.048 & 0.074 & 0.035 \\
                    &                     & 4 & 0.066  & 0.029 & 0.046 & 0.035  & 0.043 & 0.062 & 0.053 \\
\hline 
\multirow{6}{*}{CS} & \multirow{3}{*}{25} & 2 & 0.065  & 0.036 & 0.041 & 0.055  & 0.037 & 0.058 & 0.05  \\
                    &                     & 3 & 0.058  & 0.036 & 0.041 & 0.042  & 0.039 & 0.058 & 0.04  \\
                    &                     & 4 & 0.048  & 0.033 & 0.032 & 0.051  & 0.032 & 0.053 & 0.046 \\
\cline{2-10}
                    & \multirow{3}{*}{50} & 2 & 0.064  & 0.032 & 0.041 & 0.044  & 0.030  & 0.067 & 0.047 \\
                    &                     & 3 & 0.056  & 0.026 & 0.028 & 0.051  & 0.048 & 0.074 & 0.040  \\
                    &                     & 4 & 0.066  & 0.029 & 0.046 & 0.035  & 0.043 & 0.062 & 0.054 \\
\hline
\hline 
\end{tabular}
\end{table*}

\noindent \textbf{Setup 2:}  We employed a Compound Symmetry (CS) structure to model the covariance matrix, where $\Sigma_{p,q} = \theta$ if $p \neq 1$, and $\Sigma_{p,q} = \theta$ if $p = q$. This CS structure simulates a scenario in which all words within a document share the same correlation. For power comparison, $\theta_A$ and $\theta_B$ represent the parameter $\theta$ for group A and group B, respectively. Specifically, $\theta_A$ was set to $0.5 + \delta/2$ and $\theta_B$ to $0.5 - \delta/2$, with $\delta \in \{0.1,0.2,0.3,0.4\}$ and the parameter $S \in \{2,3,4\}$. This setup assesses the performance of the proposed methods under various conditions. For size comparison, we set $\theta_A = \theta_B = 0.5$ and varied $S \in \{2,3,4\}$. The results are reported in Table~\ref{SIMU1Power} and Table~\ref{SIMU12Size}. The effectiveness of the proposed method in detecting pattern differences is validated through simulations across both Setup 1 and Setup 2.  

\begin{table*}[ht]
\footnotesize
\caption{Power comparison under CS type covariance matrix setup}
\centering
\small
\begin{tabular}{c|c|c|cccc|ccc}
\hline
\hline
Param  S                  & Param V              & Param $\delta$ & Cauchy & MPT1  & MPT2  & Single & LDA   & C2ST  & MMD   \\
\hline 
\multirow{8}{*}{4} & \multirow{4}{*}{25}  
& 0.1 &    0.622 & 0.555 & 0.569 & 0.249 & 0.095
& 0.131 & 0.095 \\ 
& & 0.2    & 0.780   & 0.723 & 0.738 & 0.684  & 0.53  & 0.431 & 0.354 \\
                   &                   & 0.3    & 0.858  & 0.828 & 0.827 & 0.792  & 0.739 & 0.491 & 0.539 \\
                   &                   & 0.4    & 0.912  & 0.890  & 0.894 & 0.860   & 0.863 & 0.550  & 0.740  \\
                    \cline{2-10}
                   & \multirow{4}{*}{50} & 0.1    & 0.602  & 0.534 & 0.548 & 0.478  & 0.089 & 0.222 & 0.15  \\ 
                   &                     & 0.2    & 0.774  & 0.717 & 0.737 & 0.681  & 0.368 & 0.392 & 0.344 \\
                   &                     & 0.3    & 0.844  & 0.811 & 0.821 & 0.762  & 0.622 & 0.455 & 0.538 \\
                   &                     & 0.4    & 0.892  & 0.872 & 0.874 & 0.845  & 0.793 & 0.494 & 0.745 \\
\hline 
\multirow{8}{*}{3} & \multirow{4}{*}{25} & 0.1    & 0.521  & 0.423 & 0.445 & 0.386  & 0.111 & 0.157 & 0.123 \\
                   &                     & 0.2    & 0.711  & 0.651 & 0.666 & 0.601  & 0.442 & 0.334 & 0.299 \\
                   &                     & 0.3    & 0.808  & 0.767 & 0.778 & 0.743  & 0.688 & 0.424 & 0.462 \\
                   &                     & 0.4    & 0.858  & 0.827 & 0.831 & 0.783  & 0.813 & 0.460  & 0.64  \\ \cline{2-10}
                   & \multirow{4}{*}{50} & 0.1    & 0.496  & 0.418 & 0.431 & 0.379  & 0.079 & 0.173 & 0.113 \\
                   &                     & 0.2    & 0.703  & 0.649 & 0.667 & 0.602  & 0.284 & 0.311 & 0.317 \\
                   &                     & 0.3    & 0.792  & 0.741 & 0.753 & 0.696  & 0.551 & 0.387 & 0.469 \\
                   &                     & 0.4    & 0.862  & 0.826 & 0.829 & 0.756  & 0.702 & 0.437 & 0.661 \\
\hline 
\multirow{8}{*}{2} & \multirow{4}{*}{25} & 0.1    & 0.364  & 0.292 & 0.308 & 0.267  & 0.071 & 0.088 & 0.087 \\
                   &                     & 0.2    & 0.557  & 0.502 & 0.505 & 0.453  & 0.239 & 0.199 & 0.232 \\
                   &                     & 0.3    & 0.659  & 0.601 & 0.611 & 0.556  & 0.466 & 0.298 & 0.39  \\
                   &                     & 0.4    & 0.718  & 0.665 & 0.67  & 0.619  & 0.650  & 0.332 & 0.530  \\ \cline{2-10}
                   & \multirow{4}{*}{50} & 0.1    & 0.363  & 0.289 & 0.306 & 0.252  & 0.058 & 0.087 & 0.07  \\
                   &                     & 0.2    & 0.522  & 0.459 & 0.475 & 0.429  & 0.147 & 0.177 & 0.227 \\
                   &                     & 0.3    & 0.615  & 0.564 & 0.578 & 0.522  & 0.327 & 0.256 & 0.399 \\
                   &                     & 0.4    & 0.681  & 0.642 & 0.651 & 0.593  & 0.489 & 0.316 & 0.532 \\
\hline
\hline
\end{tabular}
\end{table*}

\subsection{Benchmark dataset}
\label{subsec: benchmark data}
We evaluate the performance of the proposed test using the AG News dataset, a well-known benchmark in the field of natural language processing \citep{sachan2019revisiting, yang2019xlnet}. The dataset contains news articles categorized into four topics: \textit{World} News (C1), \textit{Sports} News (C2), \textit{Business} News (C3), and \textit{Sci/Tech} News (C4), with each category containing 30,000 news articles. 

In the experiment, we randomly select $N$ samples from a specific category, followed by a second random selection of $N$ samples from the same category. For example, we first select $N$ articles from the World News category (C1) and then select another $N$ article from C1. After obtaining these samples, we apply the proposed method and the competing methods, each at the significance level $\alpha = 0.05$, and compare the test results. The experiments are carried out for $N = 100$ and $N = 500$.
\begin{table*}[!ht]
\caption{\label{tab:Simulation1_Size} Test size comparison on AG News benchmark data with different text generation configurations}
\small
\centering
\begin{tabular}{c|c|c|c|c|c}
\hline 
\hline
Sample size & Method & C1 vs C1 & C2 vs C2 & C3 vs C3 & C4 vs C4 \\ \hline
\multirow{7}{*}{N = 100} & {Single} & 0.051 & 0.053 & 0.060 & 0.058 \\ 
 & Cauchy & 0.060 & 0.059 & 0.073 & 0.068 \\
 & MPT1 & 0.021 & 0.031 & 0.024 & 0.027 \\ 
 & MPT2 & 0.026 & 0.024 & 0.026 & 0.025 \\ \cline{2-6} 
 & LDA & 0.041 & 0.049 & 0.044 & 0.057 \\ 
 & MMD & 0.040 & 0.051 & 0.041 & 0.043 \\ 
 & C2ST & 0.055 & 0.067 & 0.072 & 0.066 \\ \hline
\multirow{7}{*}{N = 500} & {Single} & 0.058 & 0.070 & 0.065 & 0.055 \\ 
 & Cauchy & 0.080 & 0.099 & 0.081 & 0.079 \\
 & MPT1 & 0.041 & 0.054 & 0.036 & 0.033 \\
 & MPT2 & 0.037 & 0.048 & 0.037 & 0.031 \\ \cline{2-6} 
 & LDA & 0.039 & 0.044 & 0.042 & 0.042 \\
 & MMD & 0.000 & 0.000 & 0.000 & 0.000 \\ 
 & C2ST & 0.063 & 0.069 & 0.054 & 0.057 \\ 
 \hline 
 \hline
\end{tabular}
\end{table*}

\begin{table*}[!ht]
\caption{\label{tab:Simulation1_Power}Power comparison on AG News benchmark data with different text generation configurations}
\footnotesize
\centering
\begin{tabular}{c|c|c|c|c|c|c|c}
\hline
\hline
Sample size & Method & C1 vs C2 & C1 vs C3 & C1 vs C4 & C2 vs C3 & C2 vs C4 & C3 vs C4 \\ \hline
\multirow{7}{*}{N = 100} & Single & 0.943 & 0.710 & 0.882 & 0.870 & 0.941 & 0.857 \\ 
 & Cauchy & 0.999 & 0.904 & 0.995 & 0.984 & 0.999 & 0.987 \\ 
 & MPT1 & 0.980 & 0.802 & 0.979 & 0.943 & 0.988 & 0.938 \\
 & MPT2 & 0.972 & 0.784 & 0.961 & 0.944 & 0.978 & 0.934 \\ 
\cline{2-8} 
 & LDA & 0.408 & 0.173 & 0.199 & 0.406 & 0.195 & 0.171 \\
 & MMD & 0.426 & 0.253 & 0.620 & 0.565 & 0.583 & 0.708 \\
 & C2ST & 0.132 & 0.157 & 0.116 & 0.234 & 0.152 & 0.096 \\ \hline
\multirow{7}{*}{N = 500} & Single & 1.000 & 1.000 & 1.000 & 1.000 & 1.000 & 1.000 \\ 
 & Cauchy & 1.000 & 1.000 & 1.000 & 1.000 & 1.000 & 1.000 \\ 
 & MPT1 & 1.000 & 1.000 & 1.000 & 1.000 & 1.000 & 1.000 \\ 
 & MPT2 & 1.000 & 1.000 & 1.000 & 1.000 & 1.000 & 1.000 \\ \cline{2-8} 
 & LDA & 0.995 & 0.835 & 0.870 & 0.998 & 0.858 & 0.814 \\
 & MMD & 0.521 & 0.453 & 0.605 & 0.748 & 0.709 & 0.782 \\
 & C2ST & 0.810 & 0.773 & 0.742 & 0.921 & 0.746 & 0.885 \\ \hline
 \hline
\end{tabular}
\end{table*}
For power comparison, we first randomly select $N$ samples from one category, followed by another $N$ samples from a different category. For example, in a comparison between C1 and C2, the first batch of $N$ articles is drawn from the category C1, and the second batch is sourced from C2. Table~\ref{tab:Simulation1_Power} presents the results of the comparison. 

At a sample size of $N =100$, our method outperforms competing techniques, achieving an average power of 0.8 in six data generation configurations, compared to an average power of approximately 0.25 for other methods. When the sample size is increased to $N=500$, our method maintains its superior performance, consistently achieving the highest power in all scenarios tested. These results highlight the effectiveness of our proposed method in detecting differences between text samples, regardless of sample sizes. 

\subsection{Real data example}
\label{subsec: US crash data}
Advanced Driver Assistance Systems (ADAS) have the potential to reduce traffic crashes and improve safety \citep{bengler2014three}. The U.S. National Highway Traffic Safety Administration (NHTSA)  has been collecting crash data since the early 1970s through various programs, with the Crash Investigation Sampling System (CISS) being one of the largest crash databases in the world \citep{NHTSACISS}. These crash reports include detailed narratives written by police officers or trained investigators describing the factors that contribute to each crash. 

Our analysis investigates differences in crash narratives in various factors, focusing on comparing vehicles with ADAS with those without ADAS. Table~\ref{tab: CISS_ADAS} provides an overview of the ADAS categories and the corresponding CISS cases by subgroup. NHTSA classified the ADAS into three categories: Collision Warning, Collision Intervention, and Driving Control Assistance. Detailed descriptions of each subgroup are available from NHTSA to provide insight into how these technologies operate \citep{NHTSAADS}. Crash narratives are documented by trained investigators or law enforcement officers at the crash scene, capturing evidence and observations. Since these crash narratives follow a standardized format,  the observed text dissimilarity arises not from variations in word choice, but from differences in the underlying crash causation mechanisms.

\begin{table}[ht]
\centering
\caption{Number of U.S. CISS crash narrative samples by types of ADAS technology}
\footnotesize
\begin{tabular}{c|c}
\hline
\hline
ADAS type & Subgroup \\ \hline
 \multirow{2}{*}{Collision intervention} & Automatic emergency braking\\
 & Pedestrian automatic emergency braking \\ \hline
\multirow{2}{*}{Driver control assistance} & Adaptive cruise control  \\
 & Lane keeping support  \\ \hline
\multirow{4}{*}{Collision warning} & Lane departure warning  \\
 & Forward collision warning  \\
 & Blind spot detection \\
 & Automatic crash notification \\ \hline \hline
\end{tabular}
\label{tab: CISS_ADAS}
\end{table}


Manual feature extraction is impractical for analyzing the large volume of crash narratives \citep{vallmuur2015machine}. Extracting meaningful insights from unstructured text data requires efficient statistical analysis methods capable of identifying incident patterns, which can guide safety management decisions \citep{guo2019statistical, kwayu2021discovering}.  To determine whether the observed textual dissimilarities are statistically significant, we apply the proposed method to NHTSA's CISS database, focusing on crash cases from 2018 to 2022.  We stratified the crash narratives by key risk factor subgroups. The posted speed limit is the maximum legal speed at which vehicles can travel in a particular area under ideal conditions. Road alignment refers to the curvature of a road. Road surface condition is a critical factor for vehicle control and effective braking. Adverse conditions, such as wet surfaces, ice, or snow, reduce traction and increase the risk of skidding and accidents. The post-crash integrity loss assesses the structural condition of the vehicle after the incident. A vehicle without loss of integrity has maintained its structural integrity after a crash. Pre-impact stability examines the stability of the vehicle immediately before the crash.

Table~\ref{tab: test result} presents the results of the two-sample text test, categorizing ADAS features into three main types: Intervention, Assistance, and Warning. To compare the narratives of vehicles equipped with ADAS with those without, we applied the proposed two-sample text test with Bonferroni correction to account for multiple comparisons within the same subgroup across different ADAS configurations. A checkmark in the table indicates a statistically significant difference in crash outcomes between vehicles with and without ADAS, suggesting that the presence of ADAS may influence crash dynamics. This analysis provides further insight into the key factors that affect the functionality of ADAS, vehicle safety, and crash outcomes. 

ADAS features were found to be significant under certain conditions, particularly in high-risk environments such as high speeds, curved roads, adverse weather conditions, and poor surface conditions. Significant differences were observed between different environments and crash severity levels for the intervention feature. The key factors analyzed included the posted speed limit, alignment of the roadway, condition of the road surface, and lighting. The intervention feature was significantly effective in scenarios involving speeds above 55 mph, curved roads, poor road surfaces, severe crashes, loss of post-crash vehicle integrity, and lack of pre-impact stability. In contrast, the warning and assistance features did not show significant outcomes in these conditions. Lighting conditions, representing natural or artificial illumination, did not significantly impact the performance of ADAS features, with no difference observed between nighttime and daylight conditions for any of the three types of ADAS. However, the assistance feature demonstrated a significant effect for distracted drivers, highlighting its potential in safety enhancement in these scenarios.

\begin{table*}[!ht]
\centering
\caption{Two-Sample Text Test Results for ADAS Features Based on Environmental, Driver, and Crash Severity Factors}
\footnotesize
\begin{tabular}{c|cc|ccc}
\hline \hline
Factor                & \multicolumn{2}{c|}{Subgroup}                                                            & Intervention & Assistance & Warning \\ \hline
\multirow{11}{*}{Environmental} & \multicolumn{1}{c|}{\multirow{3}{*}{Posted speed limit}}             & $\leq$ 40 MPH & $\times$              & $\times$            & $\times$         \\
                                & \multicolumn{1}{c|}{}                                                & 40-55 MPH                  & $\times$              & $\times$            & $\times$         \\
                                & \multicolumn{1}{c|}{}                                                & $\geq$ 55 MPH & $\checkmark$          & $\times$            & $\times$         \\ \cline{2-6} 
                                & \multicolumn{1}{c|}{\multirow{2}{*}{Roadway   alignment}}            & Straight                   & $\times$              & $\times$            & $\times$         \\
                                & \multicolumn{1}{c|}{}                                                & Curved                     & $\checkmark$          & $\times$            & $\times$         \\ \cline{2-6} 
                                & \multicolumn{1}{c|}{\multirow{2}{*}{Surface   condition}}            & Dry                        & $\times$              & $\times$            & $\times$         \\
                                & \multicolumn{1}{c|}{}                                                & Bad                        & $\checkmark$          & $\times$            & $\times$         \\ \cline{2-6} 
                                & \multicolumn{1}{c|}{\multirow{2}{*}{Light   condition}}              & Night                      & $\times$              & $\times$            & $\times$         \\
                                & \multicolumn{1}{c|}{}                                                & Daylight                   & $\times$              & $\times$            & $\times$         \\ \hline
\multirow{2}{*}{Human factor}   & \multicolumn{1}{c|}{\multirow{2}{*}{Driver distraction}}             & Distracted                 & $\times$              & $\checkmark$        & $\times$         \\
                                & \multicolumn{1}{c|}{}                                                & Attentive                  & $\times$              & $\times$            & $\times$         \\  \hline
\multirow{6}{*}{Crash severity} & \multicolumn{1}{c|}{\multirow{2}{*}{Crash severity}}                 & Severe                     & $\checkmark$          & $\times$            & $\times$         \\
                                & \multicolumn{1}{c|}{}                                                & Moderate                   & $\times$              & $\times$            & $\times$         \\ \cline{2-6} 
                                & \multicolumn{1}{c|}{\multirow{2}{*}{Post   crash integrity loss}}    & No                         & $\times$              & $\times$            & $\times$         \\
                                & \multicolumn{1}{c|}{}                                                & Yes                        & $\checkmark$          & $\times$            & $\times$         \\ \cline{2-6} 
                                & \multicolumn{1}{c|}{\multirow{2}{*}{Pre-impact   stability}}         & Not stable                 & $\checkmark$          & $\times$            & $\times$         \\
                                & \multicolumn{1}{c|}{}                                                & Stable                     & $\times$              & $\times$            & $\times$         \\ \hline \hline
\end{tabular}
\label{tab: test result}
\end{table*}

\section{Discussion}
\label{sec: conclusion}
The study introduced a two-sample text test to evaluate text similarity, addressing challenges in the misalignment between word embeddings and learning objectives, the high computational complexity, and the lack of theoretical frameworks for statistical inference of textual data. The proposed test infers text similarity using the entropy of text corpora fitted by neural autoregressive language models. To mitigate power loss, a multiple data split strategy was employed, and the resulting \textit{p}-values are against a unified threshold. The effectiveness of the proposed method was validated through numerical simulations, a benchmark data set example, and the U.S. national crash report database case study. The results not only preserved the Type I error rate but also achieved superior performance compared to state-of-the-art tests in terms of power. 

Future research could explore strategies for reusing the estimation data that is not utilized in the inference phase to enhance statistical power while still maintaining the Type I error with proper guarantees. Another promising direction is to develop a test for scenarios where the autoregressive condition does not hold. It would be valuable to investigate whether the proposed test statistics can be adapted for non-autoregressive neural network structures to examine their theoretical properties in such settings.

\bibliographystyle{ieeetr}
\bibliography{references}

\end{document}